\documentclass[conference]{IEEEtran}
\IEEEoverridecommandlockouts

\usepackage[T1]{fontenc}
\usepackage{amsmath,amssymb,amsfonts}
\usepackage{graphicx}
\usepackage{booktabs}
\usepackage{multirow}
\usepackage{xcolor}
\usepackage[colorlinks=true,allcolors=blue!60!black]{hyperref}
\usepackage[capitalise]{cleveref}
\usepackage{stfloats}
\usepackage{balance}
\usepackage{listings}
\usepackage{etoolbox}
\newcommand{\tablefont}{\scriptsize}
\AtBeginEnvironment{table}{\tablefont\setlength{\tabcolsep}{4pt}}
\AtBeginEnvironment{table*}{\tablefont\setlength{\tabcolsep}{4pt}}
\newcommand{\Cr}{C_{\mathrm{r}}}
\newcommand{\Cw}{C_{\mathrm{w}}}
\newcommand{\Ca}{C_{\mathrm{a}}}
\newcommand{\Cs}{C_{\mathrm{sense}}}
\newcommand{\EVact}{\mathrm{EV}_{\mathrm{act}}}
\newcommand{\EVask}{\mathrm{EV}_{\mathrm{ask}}}

\title{When Should a Failing Robot Ask?\\
Initiating Corrective Human-Robot Dialogue from Audited Sensor Evidence}

\author{
\IEEEauthorblockN{Eshika Pathak\IEEEauthorrefmark{1}\IEEEauthorrefmark{2} and Leela Krishna\IEEEauthorrefmark{2}}
\IEEEauthorblockA{\IEEEauthorrefmark{1}University of Illinois Urbana-Champaign \quad \IEEEauthorrefmark{2}Centific}
\thanks{\IEEEauthorrefmark{1}Work done while at Centific. Email: epathak2@illinois.edu}
}

\begin{document}
\maketitle

\begin{abstract}
A robot that fails at a task faces the first decision in corrective dialogue: act on its own diagnosis, consult another onboard sensor, or interrupt a person. We study when initiating dialogue is necessary relative to the evidence the acting model actually has. Choosing well requires two things current systems lack: knowing how much the robot's sensors reveal about the cause, and knowing how reliable the robot's own diagnosis is. We build a simulated benchmark in which every failure's true cause is known, because we injected it, and measure what each sensor reveals, with explicit checks against data leakage. Some failures are diagnosable from camera images; others only from the robot's force data (0.99 from force data, no image method above 0.55). We then test six open vision-language models. Their behavior tracks the surface of the prompt, not the evidence: moving the refusal option from last to first in the answer list collapses refusal rates from 78--100\% to 0--6\% in three of the six swept model-and-family pairs. Both Cosmos generations' refusal on grasp failures survives every variant. Accuracy from frames stays at or below a majority-class baseline under every prompt variant, with or without worked examples, and stated confidence carries no information about correctness. Handing the same models the force data as ten lines of text produces the study's first above-baseline diagnoses, in four of the six models: much of the failure reflects missing sensor data, not missing ability. We therefore pose the choice as a three-action decision problem, act, consult your own sensors, or ask a human, whose optimal policy follows from measured accuracy. The models do not follow it, and their ask rates ignore a fourfold change in question cost. One question to a human still lifts them from that baseline to roughly the answerer's own reliability (0.70--0.81 when they ask). Today, the decision to ask should be wired to measured accuracy and stated costs, not to the model's confidence.
\end{abstract}

\section{Introduction}
\label{sec:intro}

\looseness=-1 Corrective human-robot dialogue begins before either party has said anything. When a robot working alongside a person fails~\cite{honig}, its first communicative decision is whether to keep acting on its own diagnosis or interrupt the task and ask. A question costs attention, delays the task, and may change whether the physical state is still recoverable. Detecting that something went wrong is well studied~\cite{codeasmonitor}, and repair is straightforward once the cause is known; the unexamined step sits between them, and it is a dialogue decision: when should the robot stop acting and start talking?

\looseness=-1 By dialogue we mean situated corrective guidance after a failed attempt, in which the robot may ask what went wrong, confirm a suspected cause, or request a correction, and the person may answer, demonstrate, or take the arm. Our experiments isolate the first move of that exchange, whether to initiate it at all, and the comprehension step that follows.

\looseness=-1 The literature holds two opposite assumptions about this diagnosis stage. One line of work names the cause of a failure from the robot's observations, either by reasoning over textual summaries of an episode without training~\cite{reflect} or by training on failures synthesized from successful demonstrations~\cite{aha,pacaud2025scaling,dream2fix}, with real-world failure benchmarks following~\cite{vifailback}. What these share is the assumption that the cause is recoverable from the observations given; none measures whether it is. A second line builds mechanisms for asking humans for help~\cite{knowno,tellex-inverse,inteliplan}, assuming that sometimes it is not; neither line checks which assumption holds for a given failure. We built a benchmark to check; the answer is that diagnosability depends on which failure it is and which sensor you consult.

\looseness=-1 The study answers three questions in order. First, \emph{what can be known?} We create failures with known causes and measure how well each sensor reveals the cause, with classifiers screened for data leakage (\cref{sec:audit}). Second, \emph{do vision-language models know it?} We test whether six open models recover the recoverable causes and whether their confidence reflects what can be known (\cref{sec:results-gap}). Third, \emph{do they choose well?} We give the models the act-or-ask choice under explicit costs and compare their choices to the mathematically best policy (\cref{sec:results-decision}). One quantity links the three: whether to ask depends on $p$, the probability that the robot's own diagnosis is correct (\cref{sec:decision-math}). The audit bounds $p$ by what the sensor data contain, the diagnosis test measures what each model extracts, and with both measured the best choice is computable and each model is scored against it.

\looseness=-1 The sensor measurements were more specific than we expected. Detection failures (the robot cannot find the object it was asked to fetch) can be diagnosed from images, and the score survives changes of viewpoint and scene appearance. Grasp failures (the object slips out during a lift) can be diagnosed almost perfectly from the robot's force data (wrist force-torque and gripper measurements), at 0.99, but not from its camera: a sequence of increasingly capable image classifiers, ending with a pretrained vision backbone and a video model trained end to end, levels off at 0.55. The cause of a grasp failure is recorded inside the robot and largely invisible from outside, so a frames-only diagnoser cannot reliably tell what happened even though the robot's own sensors could settle it; whether human observers do better is untested here. A third family, placement failures, shows the pitfall of this kind of measurement. Its image classifier scores 0.998, but the score does not come from evidence about the failure: one injected cause moves the container, so the classifier succeeds by learning where the container sits in the frame, and from camera angles held out of training its accuracy falls to 0.605. We keep this family as the concrete case for why no score is believed here until it survives such tests (\cref{sec:audit}). We also state plainly which part of the grasp-versus-detection contrast follows from our own construction: grasp causes are physical parameters the renderer never draws (grip force, object mass, surface friction), while detection causes are visible properties of the scene (an occluder, an absent object, dim lighting), so images were always going to reveal more about detection failures than about grasp failures. What the construction does not determine, the experiments do: the actual amount of information each sensor carries, and whether a given high score is genuine, which the detection score proves by surviving the held-out tests and the placement score fails.

\looseness=-1 The six models fail in a way we did not anticipate. At first sight they split into two camps: three refuse to diagnose nearly everything, three commit to a cause on nearly every episode, and both camps score at or below a majority-class baseline, which always guesses the most common cause. A robustness sweep dissolves the split. Moving the ``cannot be determined'' option from the last position in the list to the first collapses the refusals almost entirely in three of the six model-and-family pairs we swept, and one model then picks whatever option sits second on 33 of 35 episodes regardless of content. The models are not cautious or reckless; they are answering the layout of the prompt. Accuracy from frames stays at or below the majority-class baseline under every variant, and stated confidence carries no information in any of them. One behavior survives the sweep: the refusal on grasp failures persists across all six variants for both Cosmos models. What no model does, under any prompt we tried, is behave differently between failures whose cause is demonstrably recoverable from its input and failures whose cause is not.

\looseness=-1 Worked examples in the prompt move no model toward what the images verifiably contain, so the failure is not a zero-shot artifact. Handing the models the robot's force data as a few lines of text does move most of them: four of the six then diagnose above the majority-class baseline, the first above-baseline results in this study. And when the models ask a human, the single answer does almost all the work: the four models with decision-task ask data rise from at or below the majority-class baseline to 0.70--0.81 when they ask, roughly the answerer's 0.80 reliability passing through.

\looseness=-1 This paper contributes:
\begin{enumerate}\setlength{\itemsep}{0pt}\setlength{\parskip}{0pt}\setlength{\topsep}{1pt}
\item Audited measurements of which failures are diagnosable from which sensors, including one family where the camera is nearly useless and force data nearly sufficient (\cref{sec:audit}).
\item Evidence from six open 7B--16B vision-language models that diagnosis behavior follows prompt layout rather than evidence, with accuracy from frames at or below a majority-class baseline under every variant and no informative confidence channel but one (\cref{sec:results-gap}).
\item A way to decide among acting, sensing, and asking: explicit costs plus measured accuracy give an optimal policy, and a model's distance from it is reported as wasted cost (\cref{sec:decision-math}).
\item The measured value and misuse of asking: one question lifts models from at or below the majority-class baseline to roughly the answerer's reliability (0.70--0.81 when they ask), yet no ask rate tracks the cost of asking (\cref{sec:results-decision}).
\end{enumerate}

\section{Related Work}
\label{sec:related}

\looseness=-1 \textbf{Diagnosing failures.} Systems name the cause of a failure from robot observations, by prompting a language model over textual episode summaries~\cite{reflect}, by training on synthesized failures~\cite{aha,pacaud2025scaling,dream2fix,racer}, or from real-world trajectories~\cite{vifailback}. All assume the cause is recoverable from the observations given, and none verifies that assumption against data leakage; our third family shows what verification catches.

\looseness=-1 \textbf{Corrective dialogue and asking for help.} A robot's turn in a shared task updates both the conversation and the physical state, so managing it involves grounding, clarification, and repair~\cite{reimann2024survey,lukin2024scout}. Prior work generates help requests by modeling how a listener will interpret them~\cite{tellex-inverse,knepper}, recovers from faults through dialogue~\cite{blankenburg}, and learns from natural-language corrections~\cite{yay,droc,sharma2022correcting}; in the latter the human initiates, so the robot's problem is comprehension rather than initiation. Ours inverts that, and we then measure whether the reply can be used.

\looseness=-1 \textbf{When to ask, and at what cost.} KnowNo~\cite{knowno} decides when to ask using conformal prediction, giving a coverage guarantee rather than a cost-optimal policy; Ask When It Pays~\cite{askpays} prices questions in goal navigation from an information-gain analysis; ESearch-R1~\cite{esearchr1} unifies asking, memory retrieval, and navigation into one cost-aware process and learns the policy by reinforcement learning. We instead compute the reference policy from audited sensor evidence, measured model accuracy, and stated costs, which lets us score off-the-shelf models against it rather than train one. Related benchmarks examine when software agents defer to a human~\cite{hilbench}, decouple question quality from navigation~\cite{qasknav}, and document models that rarely refuse to answer~\cite{abstaineqa,yesman}; because option order alone can produce apparent refusal~\cite{zhengmcq}, we test that directly. Our selective-asking comparison follows selective classification~\cite{elyaniv}. \Cref{sec:appendix-related} expands on the closest systems.

\section{Problem Setup}
\label{sec:formulation}

\subsection{Failures with known causes}
For real failures the true cause is unobservable, so no ground truth exists to score a diagnosis against; we therefore generate failures whose causes are known by construction. Each episode samples a cause $c$ from a family-specific set $\mathcal{C}=\{c_1,\dots,c_k\}$, samples a severity from a cause-specific range, and runs a scripted pick-and-place policy in a physics simulator~\cite{libero} under that perturbation; failing episodes are retained, each labeled with its injected cause. Two observation channels are recorded per episode: the camera frames that the evaluated vision-language models receive, and the robot's proprioceptive and force measurements (wrist force-torque, gripper aperture, kinematic status), referred to below as the \emph{force data}, which the models never see in the frames-only conditions but which the classifiers of \cref{sec:audit} use.

\subsection{Measuring what a sensor reveals}
\label{sec:legibility-def}
To measure how much a sensor's data reveals about the cause, we train a small classifier to predict the injected cause from that data and treat its accuracy as a lower bound on the information present: if a simple classifier recovers the cause at 0.91, the information is demonstrably there. A high score, however, can reflect label leakage or shortcut features rather than genuine evidence about the failure. To prevent crediting a sensor with information it does not carry, we accept a score only after three checks, which together we call the audit. \emph{Shuffle check}: retrain with the cause labels randomly scrambled; accuracy must fall to chance, otherwise the label is leaking into the data through some side channel. \emph{Transfer check}: test on camera viewpoints and scene appearances held out of training; accuracy must survive, judged by a proportional rule (a drop larger than 25\% of the margin above chance fails). A classifier that scores 0.99 but collapses from a camera angle it never saw has learned the scene, not the failure; \cref{sec:audit} shows precisely this in our placement family. \emph{Separation check}: training and test episodes must come from different simulation runs, so near-duplicate episodes cannot straddle the split.

\looseness=-1 A score that passes all three is a reliable lower bound, established by exhibiting the classifier, and no future model can lower it; we call it that sensor's \emph{certificate} for the family, and it appears as \emph{Cert.}\ in the tables. The reverse claim, that a sensor does \emph{not} contain the answer, is harder, because it quantifies over every possible classifier. We support such claims only by running a sequence of checked classifiers of increasingly capable designs until their accuracies level off, and we state the result as the plateau of that sequence, never as a proven ceiling.

\subsection{The act-or-ask decision}
\label{sec:decision-math}
After a failure, an agent either \emph{acts}, executing the repair matched to its diagnosis, or \emph{asks} a human one question and then acts. Each cause has one scripted repair (for a weak grip, squeeze harder and retry; for a heavy object, change the lift), so the diagnosis selects among prepared repairs and a wrong diagnosis triggers the wrong repair. Costs are stated in units of one retry: a correct repair costs $\Cr{=}1$, a wrong repair costs $\Cw{=}10$ (execute it, undo it, start over), and a question costs $\Ca{=}3$ (a person's interrupted attention). The human's answer is imperfect, for reasons given in \cref{sec:oracle}: correct with probability $q_c{=}0.80$, uninformative with $q_v{=}0.15$, wrong with $q_w{=}0.05$. Let $p$ be the probability that the agent's diagnosis on an episode is correct. The expected costs of the two choices are
\begin{align}
\EVact(p) &= p\,\Cr + (1-p)(\Cw + \Cr), \label{eq:evact}\\
\EVask(p) &= \Ca + q_c \Cr + q_v\,\EVact(p) + q_w (\Cw + \Cr), \label{eq:evask}
\end{align}
and setting them equal gives the threshold
\begin{equation}
p^\star = \frac{(1-q_v)(\Cw+\Cr) - \Ca - q_c \Cr - q_w(\Cw+\Cr)}{(1-q_v)\,\Cw},
\label{eq:pstar}
\end{equation}
below which asking is the better choice. At the default costs $p^\star \approx 0.588$. A robot whose diagnosis is right 40\% of the time expects cost $11 - 10(0.4) = 7.0$ by acting and $6 - 1.5(0.4) = 5.4$ by asking, so it should ask; at 80\% accuracy the numbers are $3.0$ and $4.8$ and it should act. Sweeping $\Ca$ over $\{1,3,6,10\}$ moves $p^\star$ over $\{0.82, 0.59, 0.24, \text{never ask}\}$, so a rational agent's ask rate must move with the cost. A third action, consulting an onboard sensor at cost $\Cs$ (\cref{sec:protocol}), is evaluated by the same comparison. The threshold itself is classical, the value of information in the sense of Howard~\cite{howard1966} and the metareasoning tradition~\cite{russellwefald}. Our contribution is not the formula but its two inputs, which no prior failure benchmark supplies: audited measurements of how much each sensor reveals about each failure type (\cref{sec:audit}), and measured diagnosis accuracies for each evaluated model (\cref{sec:results-gap}), so that the optimal policy is computable and each model's behavior can be scored against it.

\looseness=-1 Note that $p$ is a property of the agent, not of the world: on a diagnosable failure a strong classifier exists, yet an agent whose own accuracy is low should still ask, because the decision depends on the agent's accuracy, not on the best achievable one. We therefore measure each model's accuracy $a$ per failure type on a held-out calibration set and compare four policies: the \emph{best fixed policy}, which knows only $a$ and asks on every episode exactly when $a < p^\star$; a \emph{confidence policy}, which asks whenever the model's confidence falls below a swept threshold~\cite{elyaniv} and can win only if confidence predicts the model's errors; an \emph{evidence policy} (grasp failures only), which asks based on per-episode decidability (\cref{sec:audit}) and so separates knowing the evidence from knowing oneself; and an \emph{oracle policy}, which asks exactly on the model's errors, uses ground truth, and serves only as an upper bound on the value of self-knowledge. A model's \emph{regret}, its realized cost minus that of the best fixed policy (hereafter the \emph{reference policy}), weights mistakes by their consequences and compares across models, failure types, and costs. The apparent circularity of a decision rule that uses measured accuracy is addressed in \cref{sec:appendix-audit}.

\section{Benchmark and Audit}
\label{sec:benchmark}

\subsection{Three kinds of failure, and why these three}
Manipulation failures arise at several stages of a task: perception (the target is not found or is misidentified), physical interaction (the grasp or transport fails), and task preconditions (the goal state cannot be reached), alongside classes out of scope here (planning errors, hardware faults, mis-specified instructions). We construct one family per covered stage, chosen so that the plausible location of diagnostic information differs across them: pre-action scene appearance for perception, contact-time forces for interaction, and scene state for preconditions. This choice lets the benchmark identify which sensor carries the diagnostic information, not only whether a model recovers it.

\looseness=-1 All episodes are tabletop pick-and-place scenes in the LIBERO simulator~\cite{libero}: a simulated seven-degree-of-freedom arm must grasp a named object from the table and carry it to a goal location, observed by an external camera (example frames: \cref{fig:scenes}). The families are referred to as A, B, and C in tables. Each is a mixture of physical causes with sampled severities; the families were constructed under the requirement that the scripted policy succeed on at least 95\% of unperturbed episodes, so failures are attributable to the injected cause. Each family contains 2{,}200 failure episodes (2{,}000 in the audit pool and 200 held out for model evaluation),
with table texture, lighting, and camera angle randomized and logged so the transfer check can hold them out. \emph{Grasp failures (A)}: the object leaves the gripper during a lift, because grip force was scaled down by $U(0.2,0.8)$, or object mass was scaled up by $U(1.5,4.0)$, or surface friction was reduced to $U(0.05,0.30)$. \emph{Detection failures (B)}: the robot cannot find the object it was asked to fetch, because the target is occluded 40--95\%, or absent, or the instruction uses a name missing from the robot's label set, or the lighting is dimmed to $U(0.1,0.4)$. \emph{Placement failures (C)}: placing the held object fails, because the container is closed, or 70--100\% full, or moved out of reach.

\subsection{What each sensor reveals}
\label{sec:audit}
The image classifiers range from gradient-boosted trees over PCA-compressed frames to frozen DINOv2 embeddings under a boosted head and a small 3D-convolutional network; the force-data classifier uses summary statistics of the force-torque and gripper traces.

\looseness=-1 \textbf{Detection failures can be diagnosed from images.} A classifier over pixel features reaches 0.910 and passes every check: shuffled labels fall to chance; accuracy holds at 0.844 on held-out appearances and 0.813 on held-out viewpoints (\cref{tab:modality}). A single pre-attempt frame already gives 0.912: an occluder, an empty table, and dim lighting are visible before the robot moves. Detection failure is a scene-understanding problem that ends in a failure.

\looseness=-1 \textbf{Grasp failures can be diagnosed from force data, not from images.} From images, the classifier sequence of \cref{sec:legibility-def} plateaus (\cref{tab:ladder}, appendix): snapshots reach 0.442 but fail the viewpoint check (0.345, chance); adding motion gives 0.503; a frozen pretrained backbone reaches 0.545 [0.522, 0.567]; a video model trained end to end adds nothing (0.531). All but the snapshot pass the checks, and a pre-attempt frame scores at chance (0.366), as it must: grip force, weight, and friction are invisible until contact.

\looseness=-1 From the force data, a classifier reaches 0.986 using only signals a physical robot publishes: wrist force-torque, gripper aperture and force, and slip onset computed from the force discontinuity. No feature reads simulator object pose or contact flags, and restricting to force-torque and gripper alone gives the same 0.986, so the score rides on the physics. It passes every check. That includes severity-band transfer, in which the classifier is trained only on mild perturbations and tested only on severe ones, and the reverse, scoring 0.95 in the worse direction; and a stricter variant in which evaluation groups are matched on perturbation size, so that severity alone cannot reveal the cause: within these groups the causes remain separable at 0.97, against a 0.35 chance level. The cause of a grasp failure is recorded in the force data and mostly missing from the camera.

\looseness=-1 The image classifier also provides a per-episode measure of ambiguity: the probability it assigns to its most likely cause, which we call the episode's \emph{decidability}. Raw classifier probabilities are typically overconfident, so we rescale them by temperature scaling~\cite{guo2017} on held-out episodes, until the stated probability matches the observed frequency of being correct. After this correction, decidability averages 0.574, close to the classifier's 0.545 accuracy; for a calibrated measure, the average stated probability must match the frequency of being right.

\looseness=-1 \textbf{Placement failures show the pitfall.} Their classifier scores 0.998, yet accuracy falls to 0.605 from held-out camera angles: one cause moves the container, so the classifier succeeds by learning where the container sits in the frame, which is exactly what a viewpoint change alters. We report it with its held-out numbers and exclude it from the claims above; why we accept the force-data score while rejecting this one is discussed in \cref{sec:appendix-audit}.

\subsection{The human's answers}
\label{sec:oracle}
When a model asks a question, a scripted stand-in for the human replies. It knows the injected cause, rolls a die seeded per episode (all models face identical draws), and returns one short sentence: with probability 0.80, a template naming the true cause (``that one is heavier than it looks''); with probability 0.15, an unhelpful reply (``hard to say from here''); with probability 0.05, the template for a wrong cause. Replies do not depend on question wording, which is recorded and described in \cref{sec:results-questions} but has no effect in this version.

\looseness=-1 The imperfection is deliberate: a perfect answerer would build the pro-dialogue conclusion into the apparatus and assume away the possibility of inaccurate human feedback. With a noisy answerer, asking carries real risk, and reaching the answerer's own 0.80 reliability means the human's knowledge passed through intact.

\section{Experimental Protocol}
\label{sec:protocol}

\looseness=-1 \textbf{Models.} Six open vision-language models: Qwen2.5-VL-7B-Instruct~\cite{qwenvl}, Cosmos-Reason2-8B, Cosmos3-Nano~\cite{cosmos3}, InternVL3-8B, Pixtral-12B, and Llama-3.2-11B-Vision-Instruct. Cosmos3-Nano's reasoning tower is built on a Qwen3-VL backbone, so it is not architecturally independent of the Qwen entry. Coverage differs by model (the tests are defined below). Qwen2.5-VL, Cosmos-Reason2, and Cosmos3-Nano ran every test, including all three prompt versions of the decision test and the option-position sweep; InternVL3, Pixtral, and Llama ran both diagnosis rounds and the decision test in its costs-given (swept over question cost) and rule-given versions. Qwen2.5-VL and Cosmos-Reason2, the two \emph{primary models}, additionally carry the detailed confidence and question-form analyses. A smaller relative, Qwen2-VL-2B, ran the same protocol and is reported in \cref{sec:appendix-models}. Llama required an isolated serving environment and two footnoted accommodations: four input frames where eight exceeded memory, and no few-shot condition (architecture-infeasible). All models received identical prompts verbatim (\cref{sec:appendix-prompts}); unparseable responses are scored as errors.

\looseness=-1 \textbf{Diagnosis test.} Each model sees eight frames from the episode, the task instruction, and a statement that the attempt failed, then answers a multiple-choice question over that failure type's causes plus an explicit ``cannot be determined'' option. The models receive frames at least as informative as those the classifiers used (same episodes, equal or higher resolution), so the certificates are a fair yardstick for what the models' input contains; we do not expect the models to match them, only their confidence to reflect the gap. Confidence is recorded three ways: a stated 0--100 rating; the probability the model itself assigned to its chosen option, read from its output distribution; and how often the same answer recurs across five runs sampled at temperature 0.7, the parameter that scales the output distribution before sampling (the committed answer itself is generated deterministically, and the repeated-run readout requires a positive temperature).

\looseness=-1 Because the first round produced constant confidence (\cref{sec:results-gap}), we ran a second round with the ``cannot be determined'' option removed, which we call the \emph{forced-answer round}. This variant was registered in writing before it was run and is reported alongside the original, never instead of it. Every table reports three numbers: accuracy counting refusals as errors, accuracy on answered episodes only, and the answer rate; any one of these alone can mislead. Throughout, accuracies and scores are on the 0--1 scale, while percentages denote shares of episodes or of a gap. Two further input conditions probe where the bottleneck lies. In the telemetry conditions, the admissible force-torque and gripper channels are serialized as a fixed ten-line text summary (peak and mean forces, aperture trajectory, slip onset), the \emph{telemetry text}, and given to the model in place of, or alongside, the frames. In the few-shot condition, three worked examples (frames plus correct answer), drawn only from calibration episodes, precede the question.

\textbf{Accuracy measurement.} Evaluation episodes are split, balanced by cause and separated by simulation run, into a calibration set (25 per failure type) used only to measure each model's accuracy $a$, and a test set (35--36) on which all decisions are scored. 

\looseness=-1 \textbf{Decision test.} On test episodes the model chooses \texttt{ACT:\,<cause>} or \texttt{ASK:\,<one question>}, and after any answer it must commit to a cause, whose scripted repair then runs. Three prompt versions separate different failures: \emph{plain} (no costs mentioned), \emph{costs-given} (costs stated, conclusion left to the model), and \emph{rule-given} (the model is told to ask exactly when it estimates its chance of being right is below the computed $p^\star$). The cost sweep $\Ca \in \{1,3,6,10\}$ runs under the costs-given version. Policies and scoring follow \cref{sec:decision-math}; the evidence policy uses per-episode decidability on grasp failures. Question texts are recorded and classified by form, with no scoring of quality. Every mean carries a bootstrap confidence interval (10{,}000 resamples), and paired comparisons use McNemar's test~\cite{mcnemar1947}, which compares paired conditions using only the episodes where they disagree. A three-action variant adds \texttt{SENSE} at cost $\Cs{=}0.5$: the model receives the telemetry text, then must commit. Sensing is asking an instrument, with answer reliability equal to that model's measured frames-plus-telemetry accuracy; the three-way thresholds follow from the same expected-cost comparison as \cref{eq:pstar}.

\section{Results}
\label{sec:results}

\subsection{What can be known: sensor audit}
\label{sec:results-legibility}
\Cref{tab:modality} compares what the camera and the force data reveal, family by family, with placement failures shown alongside the held-out numbers that disqualify them; the full image-classifier sequence appears in \cref{tab:ladder}. The audit itself is in \cref{sec:audit}.

\begin{table*}[tbp]
\centering
\caption{What each sensor reveals, by family: grasp causes are recoverable from telemetry, not from frames.}
\label{tab:modality}
\centering
\tablefont
\setlength{\tabcolsep}{4.5pt}
\begin{tabular}{llrrrrrr}
\toprule
Fam. & Observation set & Acc. & Base. & Shuffle & Tint & Azim. & Sever. \\
\midrule
A & F/T + gripper + kinematics & 0.986 & 0.347 & 0.339 & 0.974 & -- & 0.950 \\
A & F/T + gripper only & 0.986 & 0.347 & 0.337 & 0.974 & -- & 0.961 \\
A & camera (2-frame snapshot) & 0.442 & 0.347 & 0.338 & 0.445 & 0.345 & 0.382 \\
B & camera (2-frame snapshot) & 0.910 & 0.315 & 0.267 & 0.844 & 0.813 & 0.904 \\
C & camera (2-frame snapshot) & 0.998 & 0.397 & 0.360 & 0.926 & 0.605 & 0.976 \\
C & camera, contrast-normalized & 1.000 & 0.397 & 0.352 & 0.881 & 0.683 & -- \\
\bottomrule
\end{tabular}
\par\smallskip
{\tablefont Families: A grasp, B detection, C placement. \emph{Base.}: majority-class baseline. Slip onset in the telemetry sets is computed from the force discontinuity, not from object pose. \emph{Sever.}: train on one severity band, test on the other (worst direction). For telemetry rows, \emph{Tint} is a negative control, since table appearance cannot affect force traces, and camera angle does not apply. Family C's near-perfect score fails the camera-angle check and is a cautionary example, not a result.\par}

\end{table*}

\subsection{Do the models know it: frames}
\label{sec:results-gap}
The models neither recover the recoverable causes nor reflect this in their confidence. No model clears the majority-class baseline on grasp failures from frames in any round or prompt variant (best 0.343 against a 0.400 baseline), and none approaches the 0.910 certificate on detection failures (best overall 0.17, \cref{tab:gap}). At the default prompt the models appear to split into refusers and overcommitters, but the split does not survive a robustness sweep. Moving the ``cannot be determined'' option across three phrasings and two positions collapses refusal from 100\% to 3--6\% and 78\% to 0\% for Qwen (grasp, detection) and 94\% to 3\% for Cosmos-Reason2 (detection), while Cosmos3-Nano's detection refusal only dips to 61--92\%; with the refusal option first, Qwen selects whatever option sits second on 33 of 35 grasp episodes regardless of its content, consistent with the option-position bias documented in multiple-choice evaluation of language models~\cite{zhengmcq}. Token-probability margins are decisive in both directions (medians $+0.30$ to $+0.63$ for refusing when the option is last, equally decisive avoidance when it is first): the models are confidently answering the position, not the question.

\looseness=-1 One model-and-family pair is prompt-robust: both Cosmos models refuse on grasp failures at or near 100\% across all six variants (\cref{tab:sweep-cosmos}), while their detection refusals differ sharply, the 8B collapsing and the smaller Nano barely moving (\cref{sec:appendix-models}). Accuracy is unchanged by any of this, at or below the majority-class baseline everywhere, so no prompt arrangement was concealing competence. The surviving claim is narrower and sharper: under every arrangement tried, no model's willingness to answer or stated confidence responds to whether the cause is recoverable from its input, and for most pairs the apparent epistemic stance is an artifact of option order.

\looseness=-1 The two primary models ran the full protocol, so we examine them in detail. When Qwen answers on detection failures it scores 0.250, chance among four causes; Cosmos's answered-only 1.000 rests on two episodes at a 6\% answer rate. Stated confidence is a constant: Qwen says 50 on every grasp-failure episode and Cosmos says 100, including when its answer is ``cannot be determined.'' Spearman's rank correlation~\cite{spearman1904} between confidence and per-episode decidability is $\rho = -0.15$ and $-0.13$ for the two models: no relationship.

\looseness=-1 Two analyses close the last loophole, that refusing models knew the answer and refused to say it. In the forced-answer round, every refuser scores at or below the majority-class baseline (grasp: 0.31, baseline 0.400). And on the refused episodes themselves, the model's top-ranked non-refusal option, read from the token probabilities it assigned while refusing, is no better than its forced-answer accuracy in any of the four model-and-family pairs (largest difference 0.07). What the models refused to say was at or below chance. The refusals were hiding nothing. Confidence in the forced-answer round is no longer constant, but it takes only a few distinct values and does not correlate with correctness ($\rho \approx 0$). \Cref{fig:confidence} (appendix) plots all three confidence readouts against per-episode decidability for the primaries and Cosmos3-Nano: no readout is positively informative. On the paired episodes, McNemar's test confirms the forced-answer round differs from the first round on grasp failures for both primaries (eleven episodes changed from wrong to right and none the other way, $p=0.001$) and for Cosmos on detection failures ($p=0.004$); the primaries are statistically indistinguishable from each other in both rounds ($p>0.3$), so we make no claim that either is better.

\looseness=-1 Worked examples do not help. With three in-context exemplars, no model moves toward the certificate on either family; several drop, the best few-shot result anywhere is 0.333, majority-class picking, and refusal behavior is untouched (Qwen still refuses on every grasp episode). The failure on frames is a ceiling on what these models extract from images, not an artifact of zero-shot prompting. The answer format is not the obstacle either: given the scene fact as one sentence of text, the same models map it to the correct cause at 0.90--1.00, while free-form diagnosis without options falls to 0.08 and 0.00 (\cref{sec:appendix-formatcheck}). The gap is in perceiving the manipulation, not in reasoning from it.

\begin{table*}[tbp]
\centering
\caption{Diagnosis accuracy of the six models against the certificate.}
\label{tab:gap}
\centering
\tablefont
\begin{tabular}{llrrrrrr}
\toprule
Model & Fam. & Cert. & Overall & Commit & Rate & Forced & Gap \\
\midrule
Qwen2.5-VL-7B & A & 0.545 & 0.000 & -- & 0\% & 0.314 & 0.545 \\
Qwen2.5-VL-7B & B & 0.910 & 0.056 & 0.250 & 22\% & 0.167 & 0.854 \\
Cosmos-Reason2-8B & A & 0.545 & 0.000 & -- & 0\% & 0.314 & 0.545 \\
Cosmos-Reason2-8B & B & 0.910 & 0.056 & 1.000 & 6\% & 0.306 & 0.854 \\
Cosmos3-Nano & A & 0.545 & 0.000 & -- & 0\% & 0.314 & 0.545 \\
Cosmos3-Nano & B & 0.910 & 0.000 & -- & 0\% & 0.389 & 0.910 \\
Llama-3.2-11B-V & A & 0.545 & 0.286 & 0.286 & 100\% & 0.314 & 0.259 \\
Llama-3.2-11B-V & B & 0.910 & 0.194 & 0.194 & 100\% & 0.278 & 0.716 \\
InternVL3-8B & A & 0.545 & 0.314 & 0.314 & 100\% & 0.314 & 0.231 \\
InternVL3-8B & B & 0.910 & 0.167 & 0.176 & 94\% & 0.194 & 0.743 \\
Pixtral-12B & A & 0.545 & 0.257 & 0.290 & 89\% & 0.343 & 0.288 \\
Pixtral-12B & B & 0.910 & 0.139 & 0.143 & 97\% & 0.417 & 0.771 \\
\bottomrule
\end{tabular}
\par\smallskip
{\tablefont \emph{Cert.}: the certificate from \cref{tab:modality}. \emph{Overall} counts refusals as errors; \emph{Commit}: accuracy on answered episodes, at the \emph{Rate} shown; \emph{Forced}: the pre-registered round with the refusal option removed; \emph{Gap} = Cert.\ $-$ Overall. Majority-class baselines on the test split: 0.400 (A), 0.306 (B).\par}

\end{table*}

\subsection{Do the models know it: force data}
\label{sec:results-telemetry}
If frames do not carry the cause of a grasp failure, can the models use the sensor that does? The certificate of \cref{sec:audit} establishes that the cause of a grasp failure is recoverable from the force channels at 0.986. Serializing those channels as ten lines of text and giving them to the models produces the first above-baseline diagnoses in this study (\cref{tab:telemetry}): four of the six models clear both the majority-class baseline and their own frames-only score from telemetry text alone (InternVL3 0.514, Pixtral 0.486, Llama 0.543, Cosmos3-Nano 0.457, against a 0.400 baseline and 0.31 frames-only). Qwen recovers only when the text accompanies the frames (0.286 alone, 0.514 fused). Cosmos-Reason2 does not recover under either condition (0.286 in both), a notable result for a model built for physical reasoning, while Cosmos3-Nano does recover from the same text (0.457 alone, 0.486 fused). The two differ in backbone as well as in release, so we do not read this contrast as a generational or scaling effect.

\begin{table}[tbp]
\caption{Family-A diagnosis accuracy by input condition (forced-answer round). Four of the six models recover the cause from telemetry text alone; Cosmos-Reason2 recovers under neither condition, while its smaller, newer-generation relative does.}
\label{tab:telemetry}
\centering
\tablefont
\begin{tabular}{lrrr}
\toprule
Model & Frames & Telemetry & Frames+telem. \\
\midrule
Qwen2.5-VL-7B & 0.314 & 0.286 & 0.514 \\
Cosmos-Reason2-8B & 0.314 & 0.286 & 0.286 \\
Cosmos3-Nano & 0.314 & 0.457 & 0.486 \\
InternVL3-8B & 0.314 & 0.514 & 0.457 \\
Pixtral-12B & 0.343 & 0.486 & 0.457 \\
Llama-3.2-11B-V & 0.314 & 0.543 & 0.314$^{a}$ \\
\bottomrule
\end{tabular}
\par\smallskip
{\tablefont Majority-class baseline 0.400; telemetry certificate 0.986. $^{a}$Four input frames instead of eight (memory limit).\par}

\end{table}

\looseness=-1 Adding frames to the telemetry text helps some models and hurts others: it is the difference between failing and recovering for Qwen, helps Cosmos3-Nano slightly, costs InternVL3 and Pixtral a few points, and collapses Llama's best-in-study telemetry-only 0.543 back to its frames-only level. No model approaches the 0.986 certificate, the text summary being a lossy instrument, but the direction is unambiguous: much of what looked like inability to diagnose was information sitting in a sensor the models were never given, though one model cannot use that information even when it is given. This changes the dialogue reading of a grasp failure. It is not automatically a reason to interrupt someone: for four of the six models the cheaper move is to expose the robot's own force data in a form the model can read, and a person is the right source only when no usable robot channel carries the answer.

\looseness=-1 Partial competence also makes the confidence question testable for the first time: at zero accuracy, no confidence signal could have shown itself. At 0.46--0.54 accuracy, stated confidence remains uninformative and is mildly anti-calibrated (Pixtral verbalized $\rho=-0.44$; Llama sample-frequency $\rho=-0.55$): the models are more confident when wrong. The one informative channel anywhere is Pixtral's token log-probability ($\rho=+0.50$), which also recovers 56\% of the oracle selective-asking value, while its stated confidence recovers none (InternVL3's token probabilities are excluded; see \cref{sec:appendix-models}).

\subsection{Do the models choose well: act or ask}
\label{sec:results-decision}
We first report what a question is worth, then compare the models' choices with the reference policy of \cref{sec:decision-math}. For the four models with decision-task ask data, one question brings final task accuracy to 0.70--0.81 when they ask (Qwen 0.76, Cosmos-Reason2 0.70, Llama 0.76, Cosmos3-Nano 0.81), while unaided all four sit at or below the majority-class baseline. The reply names the true cause 80\% of the time, the model maps it onto a listed option, and the matching repair runs; the shortfall from 0.80 is the vague and wrong replies. Realized competence is almost entirely the answerer's reliability passing through the model, which raises the question of whether the models deploy so valuable a resource sensibly.

\begin{figure}[tbp]
\centering
\includegraphics[width=0.5\columnwidth]{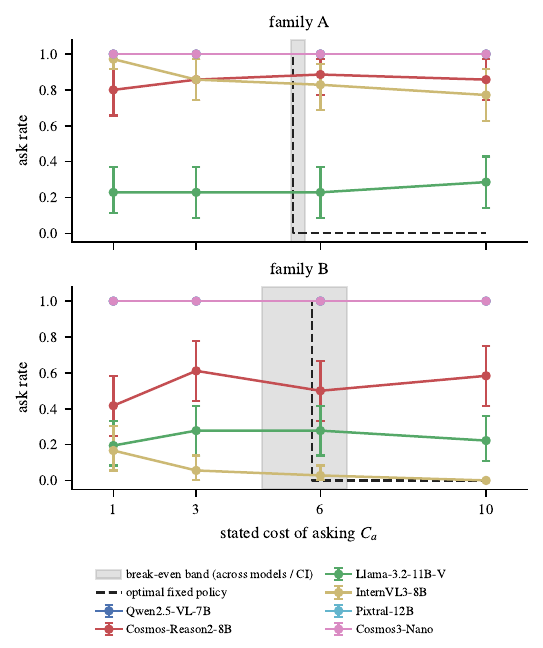}
\caption{How often each model asks as the stated cost of a question rises from 1 to 10. The dashed step is the reference policy at each model's measured forced-answer accuracy: ask until the break-even cost near 5.3, then act; the shaded band is the break-even across the accuracy estimate's interval. No model tracks the cost.}
\label{fig:askrate}
\end{figure}

\looseness=-1 With forced-answer accuracy $a \approx 0.32$ on grasp failures for the primaries (Wilson interval roughly $\pm 0.10$ at $n{=}25$), the reference policy asks at costs 1 and 3 and acts at 6 and 10; the break-even sits near 5.3 (4.6--6.6 across the estimate's interval). Neither primary tracks the cost (\cref{fig:askrate}). Qwen asks on 100\% of episodes at every cost. While the reference policy asks (costs 1 and 3), Qwen's regret is still slightly positive (0 to $+1.3$), because its commitments after unhelpful or wrong answers fall short of the reference policy's; once the reference policy switches to acting Qwen keeps asking, at growing expense ($+0.6$ to $+2.1$ at cost 6, $+4.6$ to $+6.3$ at cost 10). Its constant policy fails twice over: the wrong action when questions are dear, imperfect execution while they are cheap. Cosmos's ask rate swings with prompt wording (11\% rule-given to 89\% costs-given) but not with the cost number; the rule-given version feeds the rule its constant stated confidence of 100, so Cosmos asks least exactly where asking is optimal. Qwen ignores every instruction, including the rule.

\looseness=-1 The other four models are equally insensitive to cost, each in its own direction. InternVL3's ask rate drifts (97\% to 77\%) but never approaches the act-always policy required beyond the break-even; Pixtral asks on every episode at every cost; Cosmos3-Nano likewise always asks (350 of 350 under stated costs, 94--100\% in every prompt version); Llama under-asks throughout (19--29\%), which is costliest exactly where questions are cheap and the reference policy asks, with regret reaching $+5.05$ per episode at $\Ca{=}1$. \Cref{tab:regret} gives regret by cost and prompt version.

\looseness=-1 Confidence could in principle compensate by selecting \emph{which} episodes to ask about, but ranked by its best readout it recovers almost none of the fixed-to-oracle gap: 0\% for Qwen, InternVL3 and Llama, 8\% for Cosmos-Reason2, and 56\% only for Pixtral's token log-probability (\cref{fig:coverage}, appendix). Two simple triggers, disagreement across repeated runs and more than one cause being listed, recover 0\%.

\looseness=-1 The three-action variant behaves as the measured accuracies predict. Once questions grow expensive ($\Ca\geq6$), consulting the robot's own sensors at $\Cs{=}0.5$ becomes the reference policy's choice for four of the six models (Qwen, InternVL3, Pixtral, Cosmos3-Nano), and Qwen's predicted flip from asking to sensing lands exactly at $\Ca{=}6$. For the other two, sensing is dominated for opposite reasons: Llama cannot fuse the text with its frames, and Cosmos cannot read the text at all. Reading the instrument and profiting from it are different abilities, and a deployment should measure which its model has.

\looseness=-1 \textbf{What the answer is worth depends on how it is phrased.} The values above come from an answerer that names the cause in the words of the listed options. As a first probe of how much that matters, we replay only the commit turn with the same information phrased as a person would put it, implying the cause through an observation (``it kept sliding right out of the fingers'') or describing what a bystander saw. It changes the value of a question sharply for half the models (\cref{tab:phrasing}, with the frozen reply bank and per-register detail in \cref{sec:appendix-phrasing}): Qwen falls from 0.788 to 0.538 and Cosmos3-Nano from 0.771 to 0.480 ($p<10^{-4}$, McNemar on paired episodes), while Cosmos-Reason2 and Llama hold within noise. The two that fall were near-ceiling under menu phrasing on these episodes (0.92 and 0.98), so the highest ask values here are also the most inflated by an accommodating answerer. On the episodes each model got right under menu phrasing, the robust pair retains 0.79--0.86 of them against 0.59--0.64 for the other two, so the difference is not only headroom. The failure is lexical: with the option words removed, ``only half of it is visible'' is committed as the wrong-name cause and ``it skidded out'' as the object being absent. Asking is worth much less than the headline numbers suggest unless the person answers in the robot's vocabulary, and how much less is a property of the model. This is a single-reply probe against a scripted answerer, not an evaluation of comprehension in conversation; measuring that properly is future work (\cref{sec:limitations}).

\subsection{What the models ask}
\label{sec:results-questions}
The value of asking also depends on what is asked. The recorded questions show how the models frame a request for help. Not one took the open form: neither primary ever asked ``what went wrong?'', even on episodes it had just refused to diagnose. Qwen's grasp questions split roughly 60/40 between confirming a suspected cause and ruling one out; on detection failures 97\% are confirmations, as is every Cosmos question. They rarely target the model's own uncertainty, matching the cause it is least sure of in 14\%, 3\%, and 0\% of cases: the models ask about the cause they already favor. The wording is stereotyped too, seven distinct questions across 35 grasp episodes for Qwen and seven across all 139 asks for Cosmos.

\looseness=-1 Put next to the diagnosis results this is a direct contradiction: a model that has just said the cause cannot be determined then asks a question presupposing one. Different question forms buy different information for the same cost, the modern form of the listener-modeling problem~\cite{tellex-inverse}; scoring that choice needs a protocol like~\cite{qasknav} and richer ambiguity than ours.

\section{Limitations and Future Work}
\label{sec:limitations}
We evaluated six open 7B--16B vision-language models; frontier systems are the natural next subjects, and whether they sit above the majority-class baseline from frames or at the overconfident pole that prior work documents~\cite{abstaineqa,yesman} is directly testable with our protocol. Test sets are small (35--36 episodes per failure type), though the reported effects are extreme enough to survive this. The sensor findings are properties of these families in simulation with a scripted policy: the 0.55 image plateau belongs to our classifier sequence and this renderer, whose images omit cues real cameras carry, such as the look of a heavy object or the sag of a loaded arm. Whether the split holds under a different simulator, a learned policy, or a physical robot is open; repeating the audit on real hardware is the test we would run first. Part of the modality contrast is built in, since injected grasp causes are unrendered physics while detection causes are rendered scene properties; a family whose causes span modalities by design is in progress.

\looseness=-1 The decision layer invites extension. Costs are stipulated, swept fourfold, and a prompt variable, so the decision results measure instruction-conditioned behavior; a cause-by-repair matrix in place of the scalar $\Cw$ would tie the policy to each model's confusion structure. The answerer ignores question wording, so the ask values are conditional on a question having been asked rather than on its quality. Comprehension is the clearest opening: people paraphrase, hedge, answer a different question than the one asked, and correct themselves across turns. Evaluating that needs replies conditioned on the question, multi-turn repair, and human speakers rather than templates, with comprehension scored separately from the decision to ask, as navigation benchmarks now do for question quality~\cite{qasknav}. A deployed system would also not stop at one question but would ask, interpret, confirm, act, and re-check until the task is recovered or safely abandoned (\cref{sec:appendix-dialogue}).

\section{Conclusion}
Whether a failure hides its cause depends on the sensor and on the model that reads it: detection failures are diagnosable from images, grasp failures almost perfectly from force data but not from frames, and placement failures only appeared diagnosable until the viewpoint check. The six models cannot tell these situations apart. Refusal versus commitment is set largely by prompt layout, accuracy from frames never clears the majority-class baseline, confidence carries no information, and ask rates ignore the price of a question; yet one human reply lifts them to roughly the answerer's reliability when phrased in their terms, and when the missing information sits in the robot's own sensors, consulting them first is often rational.

\looseness=-1 For human-robot dialogue, the first turn is itself a cost-sensitive information decision: act when measured competence suffices, consult an onboard sensor when that is cheaper, and ask a person when the missing information lies outside any usable robot channel. That question should open a stateful exchange, and its value depends on whether the robot understands the answer. Both are missing here, so the policy should be tied to measured accuracy and costs, not self-reported confidence (\cref{sec:recipe}).

\bibliographystyle{IEEEtran}
\bibliography{references}

\appendices
\crefalias{section}{appendix}
\section{Audit Discussion}
\label{sec:appendix-audit}
A fair question is why we accept the force-data score for grasp failures while rejecting the image score for placement failures, when both classifiers profit from the injected cause. The distinction is not what the feature measures but what survives scrutiny. The placement classifier read the container's position in the image, a reading that collapsed under a held-out camera angle and, when suppressed by normalization, resurfaced through a different channel. The force-data classifier reads the mechanical consequences of the failure through streams any real wrist sensor produces, its score is unchanged when every ground-truth-adjacent feature is removed, and it separates causes even when perturbation magnitude is matched. In an injected-cause benchmark every honest diagnosis ultimately traces the injected cause; the standard we apply is that the tracing must run through deployable sensors and survive every transfer we can construct.

Two conventions follow from this. No filter may depend on any statistic of the number being reported, and claims that information is absent are reported as the plateau of a classifier sequence, never as a proven limit.

A related detail from the placement family: normalizing image contrast does not remove its shortcut; it moves it between the appearance and viewpoint checks (0.926 falling to 0.881 on one axis, 0.605 rising only to 0.683 on the other).

\medskip\noindent\textbf{Who measures the accuracy, and when.}\phantomsection\label{sec:who-knows} A natural objection is that deciding whether to ask requires knowing the model's accuracy in advance. The objection conflates evaluation with deployment. In evaluation, the measured accuracy defines the reference policy that regret is scored against; the gap between a model's behavior and that reference policy is the reported result, so no circularity arises. In deployment, accuracy would be measured once during commissioning, on held-out failures staged before the system is fielded, in the same way conformal methods require a calibration set before their guarantees hold~\cite{knowno}. Only a model whose confidence tracked its correctness episode by episode could skip this step and adapt from the inside; \cref{sec:results-gap} shows that ability is missing in the models we test, which is why one externally measured number outperforms their own judgment.

\section{Free-Form Diagnosis Probe}
\label{sec:appendix-formatcheck}
The multiple-choice format could in principle mask competence that free elicitation would reveal. To test this, two models with different backbones, Pixtral-12B and Cosmos3-Nano, were asked on every detection-failure test episode for a free-form diagnosis with no options: ``state what you think caused the failure.'' Accuracy falls to 0.08 and 0.00, below the multiple-choice 0.417 and 0.389, so the option list surfaces perception hypotheses the models do not generate on their own; for contrast, handing either model the ground-truth scene fact as one sentence of text yields 0.90 and 1.00 through the same multiple-choice question. Free-form answers are scored by a per-cause keyword rubric that counts any mention of the injected mechanism as correct, lenient by construction, with non-matching answers reviewed by hand.

The answers themselves show what fills the gap. On episodes with the target 62--94\% occluded, Pixtral diagnoses in grasp vocabulary (``incorrect gripper positioning,'' ``did not correctly grasp''), and Cosmos3-Nano narrates motion (``moved away instead of approaching'') while asserting it sees the hidden object (``the milk carton, which was located on the table''). On dimmed episodes, neither model ever mentions lighting. The models do not produce wrong perception hypotheses; they produce no perception hypotheses, substituting a generic manipulation story for the scene in front of them. Given the fact, the models find the cause; given the pixels, they do not find the fact.

\section{Implications for Practice}
\label{sec:recipe}
Our results suggest a concrete procedure for deployments. Diagnosis accuracy can be measured during commissioning by staging failures of each type the robot will encounter, in the same way conformal methods require a calibration set before their guarantees hold. With deployment-specific costs assigned to a repair, a wrong repair, a person's interrupted attention, and a read of the robot's own sensors, the threshold of \cref{eq:pstar} then identifies, per failure type, whether acting or consulting a cheaper information source is the better default: the robot's own sensors where they carry the answer, a person where they do not. Because any externally measured accuracy goes stale, re-measurement is warranted whenever the model, the environment, or the costs change. What the results argue against is tying this decision to the model's confidence: on our benchmark, a single externally measured number outperformed the models' episode-by-episode judgment, which carried no information.

For model builders, the results suggest two priorities. The first is identifying which sensor carries the diagnostic information before scaling the vision pipeline; for our grasp failures it is the force data, and models that could not recover the cause from frames could recover it from ten lines of text. The second is calibrated self-assessment, confidence that tracks correctness: the missing capability whose value our oracle policy bounds, and the one that would let a robot set its own threshold from the inside.

\section{Extended Related Work}
\label{sec:appendix-related}
\textbf{Failure diagnosis.} REFLECT~\cite{reflect} converts an episode into text and prompts a language model to explain the failure, without training. AHA~\cite{aha} fine-tunes a VLM on synthetically perturbed demonstrations, and further systems scale related data-generation approaches~\cite{pacaud2025scaling,dream2fix}; RACER~\cite{racer} guides imitation-learned recovery with rich language annotations; recent benchmarks draw on real-world trajectories~\cite{vifailback}. Where such a system reports imperfect accuracy, the shortfall cannot be split between a weak model and undeterminable evidence without a measurement like ours.

\textbf{Interactive planners and help-seeking.} InteLiPlan~\cite{inteliplan} incorporates human interaction into a lightweight LLM-based planner for domestic robots; we supply the evaluation that says when interrupting a person is justified. Ask When It Pays~\cite{askpays} derives question costs from an information-gain analysis and penalizes each query in its success metric, so its decision reduces uncertainty about a navigation goal, while ours is a threshold from measured diagnosis accuracy. ESearch-R1~\cite{esearchr1} is closest in structure: its Ask, GetMemory, and Navigate actions parallel our ask, sense, and act, and it trains a multimodal model to trade information gain against per-action costs. Our three actions are analogous, but the policy is computed rather than learned, which is what lets us ask whether off-the-shelf models match the optimum. QAsk-Nav~\cite{qasknav} is complementary on the evaluation side, decoupling question asking from navigation so that question quality can be scored on its own.

\section{From Initiation to Continuing Dialogue}
\label{sec:appendix-dialogue}
A deployed system would not stop at one question. It would maintain a loop of asking, interpreting, confirming, acting, and re-checking until the task is recovered or safely abandoned, which raises three questions this benchmark only opens. Spoken interaction adds turn-entry timing decisions of its own~\cite{gervits2020timing}, along with recognition and grounding errors, and full-duplex speech models make continuous listening and speaking plausible~\cite{personaplex}; these enter our framework as extra noise in the answer channel and lower the value of a question below what \cref{tab:phrasing} measures. Asking also interrupts a physical process: a robot holding an unstable object may need to reach a safe checkpoint before speaking, and the reply must then be bound to the state from which execution resumes. And corrective feedback need not be verbal at all, since a person may demonstrate, guide the arm, or teleoperate briefly~\cite{pato2023}; treating such an intervention as a dialogue turn raises the further question of whether it is a one-time rescue or a reusable correction.

\section{Answer-Phrasing Bank}
\label{sec:appendix-phrasing}
\begin{table}[tbp]
\centering
\caption{Final task accuracy when the model asks, by how the human phrases the answer. The same information is delivered in each condition; only the wording changes.}
\label{tab:phrasing}
\centering
\tablefont
\begin{tabular}{lccc}
\toprule
Model & Menu-phrased & Indirect & Narrative \\
\midrule
Qwen2.5-VL-7B & 0.788 & 0.538 & 0.599 \\
Cosmos3-Nano & 0.771 & 0.480 & 0.542 \\
Cosmos-Reason2-8B & 0.707 & 0.680 & 0.627 \\
Llama-3.2-11B-V & 0.753 & 0.691 & 0.654 \\
\bottomrule
\end{tabular}
\par\smallskip
{\tablefont Accuracy on the episodes where each model chose to ask. \emph{Menu-phrased} is the oracle used elsewhere in the paper, which names the cause in the words of the answer options; \emph{Indirect} implies the cause through an observation; \emph{Narrative} describes what a bystander saw without diagnosing. Correctness of the reply is held fixed across conditions; only the wording of the informative replies changes.\par}

\end{table}
The phrasing conditions of \cref{tab:phrasing} deliver identical information in three registers. Replies were generated once per cause, screened so that no indirect or narrative reply contains a word from any cause's vocabulary, then frozen; the correct/uninformative/wrong draw and its per-episode seed are unchanged across conditions, so only the wording of the informative replies differs. Two of the fourteen sets follow.

\vspace{3pt}\noindent\textbf{Grasp failure, object heavier than expected.}
\emph{Menu-phrased:} ``that one is heavier than it looks.''
\emph{Indirect:} ``The object barely moved when the robot lifted it, it seemed like it was too much for the arm.'' / ``The arm was trembling while trying to lift the object, it didn't go very far.'' / ``I noticed the arm was straining a lot while attempting to pick up the object.''
\emph{Narrative:} ``That thing barely lifted an inch despite the robot's best efforts.'' / ``The robot's attempt to lift that object didn't get it far off the table.'' / ``The arm seemed to struggle and the object hardly left the table's surface.''

\vspace{3pt}\noindent\textbf{Detection failure, target occluded.}
\emph{Menu-phrased:} ``it's tucked behind the other thing.''
\emph{Indirect:} ``There seems to be something blocking the view of what you're trying to pick up.'' / ``Only half of the thing you're aiming for is visible.'' / ``It looks like another object is in front of the one you're reaching for.''
\emph{Narrative:} ``Another item was in the way, blocking part of the target object.'' / ``There was a second object right in front of the one the robot tried to pick up.'' / ``I could see the target object was there, but another item was in front of it.''

\vspace{3pt}\noindent The misroutings are lexical. Given the second indirect reply above, one model commits to the instruction naming an object the robot cannot recognize, and given a slippage description (``it skidded out'') another commits to the object being absent: with the option words removed, surface overlap with the wrong option outweighs the meaning of the sentence.

\section{Diagnosis Prompt}
\label{sec:appendix-prompts}
Both rounds of the diagnosis test use the templates below, identical for every model. \verb|{options}| is the lettered cause list for the family; in the refusal-allowed round the refusal line is appended to it, with the option-position sweep varying its wording across ``cannot be determined from the available observations,'' ``not determinable from these frames,'' and ``insufficient evidence to identify the cause,'' and its position between last and first. The forced-answer round omits the refusal option entirely. Templates for the few-shot, telemetry, and decision conditions follow the same structure and are described in \cref{sec:protocol}.

\vspace{2pt}\noindent\textbf{Diagnosis, refusal allowed.}
\begin{lstlisting}
You are a robot's failure analyst. The robot attempted: "{instruction}".
The attempt failed. You are shown frames from the episode in order.
Which of the following is the cause of the failure?
{options}
Answer with the single letter, then on a new line your confidence 0-100 that this letter is correct.
\end{lstlisting}
\vspace{2pt}\noindent\textbf{Diagnosis, forced answer.}
\begin{lstlisting}
You are a robot's failure analyst. The robot attempted: "{instruction}".
The attempt failed. You are shown frames from the episode in order.
Which of the following is the most likely cause of the failure?
{options}
You must choose one option even if you are uncertain. Do not decline to answer.
Answer with the single letter, then on a new line your confidence 0-100 that this letter is correct.
\end{lstlisting}

\section{Additional Results}
\label{sec:appendix-models}
\begin{figure*}[tbp]
\centering
\includegraphics[width=\textwidth]{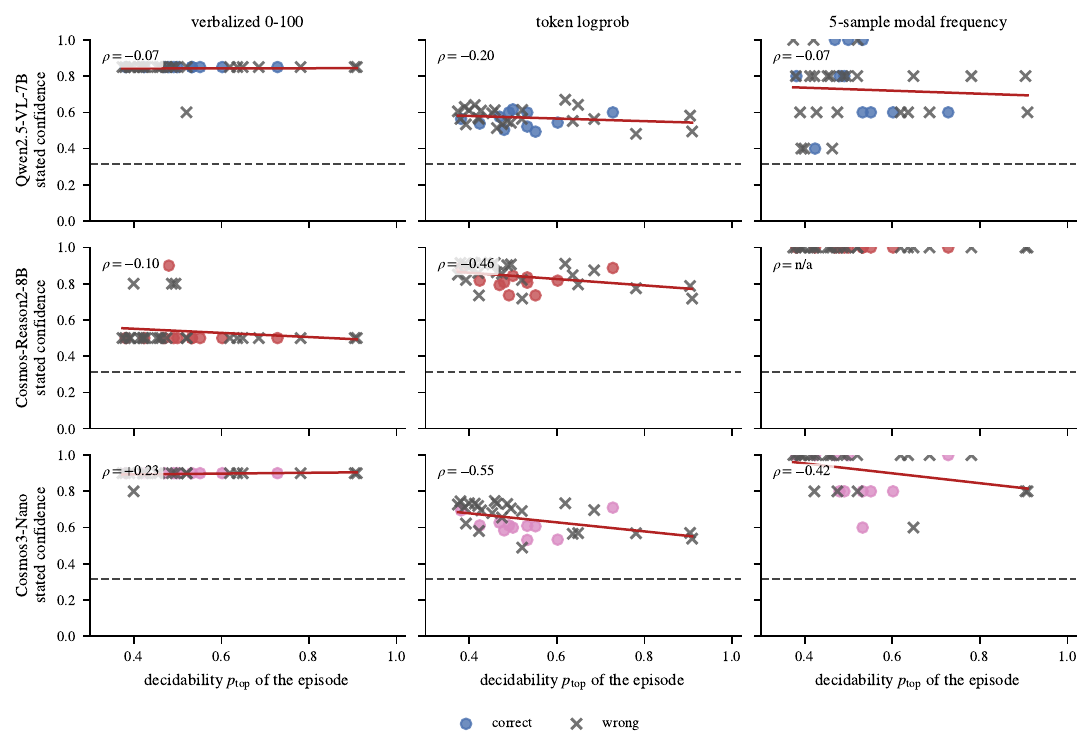}
\caption{Stated confidence does not track how much the images determine the cause. Grasp failures, forced-answer round; each column is one confidence readout; the dashed line marks the model's own accuracy (0.314 for all three). No readout is positively informative for any of the three models.}
\label{fig:confidence}
\end{figure*}

\begin{figure}[tbp]
\centering
\includegraphics[width=\columnwidth]{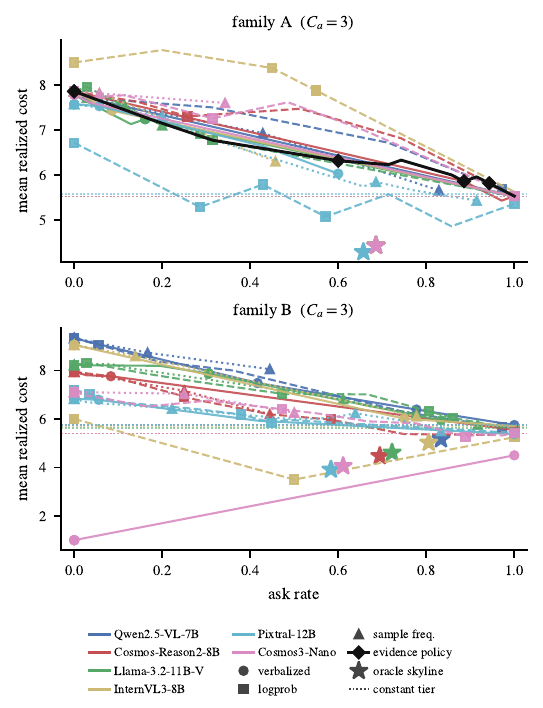}
\caption{Realized cost against ask rate at $\Ca=3$, the two family panels stacked. Curves: asking whenever a confidence readout falls below a swept threshold; the evidence policy ranks episodes by decidability instead. Dashed line: the best fixed policy (the reference policy). Star: the oracle policy that asks exactly on the model's errors (uses ground truth; an upper bound, not achievable). On the grasp panel the model curves coincide with the evidence curve: with accuracy near zero, every episode is an error, so no ranking has anything to reorder. The overlap is the result, not a rendering artifact.}
\label{fig:coverage}
\end{figure}

\begin{figure*}[p]
\centering
\includegraphics[width=\textwidth]{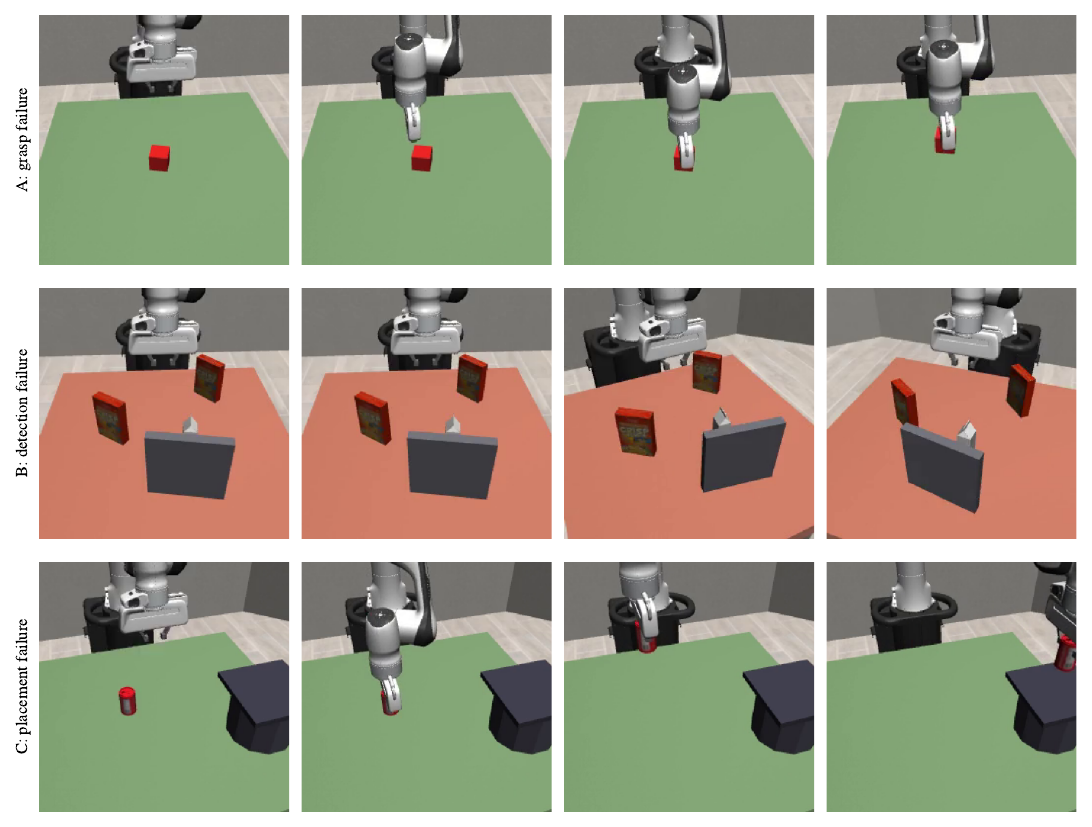}
\caption{Example episodes from the three failure families, four of the eight frames the models receive. Top (A): a grasp failure. The frames look unremarkable because the injected causes, grip force, object mass, and surface friction, are never rendered; this is why image classifiers plateau near 0.55 while the force-data classifier reaches 0.986. Middle (B): a detection failure; the occluder hiding the target is plainly visible, matching the 0.910 image certificate. Bottom (C): a placement failure; the container's state is the injected cause, and its position in the frame is the shortcut behind the disqualified 0.998 score.}
\label{fig:scenes}
\end{figure*}

\Cref{tab:gap-appendix} reports the remaining smaller relative, Qwen2-VL-2B, with the same three accuracy numbers as \cref{tab:gap}. In the decision task it rarely produces a well-formed choice, echoing both options on nearly every episode, so we do not read its decision behavior as a policy. \Cref{tab:ladder} gives the full image-classifier sequence behind the 0.55 plateau of \cref{sec:audit}. \Cref{fig:confidence} plots the three confidence readouts against decidability. \Cref{tab:regret} reports regret by model, family, prompt version, and question cost. \Cref{fig:coverage} shows the selective-asking comparison at $\Ca{=}3$. \Cref{tab:sweep-cosmos} reports the option-position sweep for both Cosmos models in full. \Cref{fig:scenes} shows example episodes from each failure family. In the confidence analysis of \cref{sec:results-telemetry}, InternVL3's token probabilities are saturated (all values within $8\times10^{-5}$ of 1.0) and carry no usable ranking; they are excluded rather than reported as a correlation.

\begin{table*}[tbp]
\centering
\caption{The remaining smaller relative, with the same three accuracy numbers as \cref{tab:gap}: at or below the majority-class baseline on both families, consistent with the main results.}
\label{tab:gap-appendix}
\centering
\small
\begin{tabular}{llrrrrrr}
\toprule
Model & Fam. & Cert. & Overall & Commit & Rate & Forced & Gap \\
\midrule
Qwen2-VL-2B & A & 0.545 & 0.000 & -- & 0\% & 0.314 & 0.545 \\
Qwen2-VL-2B & B & 0.910 & 0.167 & 0.167 & 100\% & 0.167 & 0.743 \\
\bottomrule
\end{tabular}

\end{table*}
\begin{table*}[tbp]
\centering
\caption{Image classifiers for grasp failures (family A): accuracy plateaus near 0.55.}
\label{tab:ladder}
\centering
\small
\begin{tabular}{lrrrrc}
\toprule
Image classifier & Acc. & Shuffle & Tint & Azimuth & Retained \\
\midrule
Final-state snapshot ($2{\times}24{\times}24$ px) & 0.442 & 0.338 & 0.445 & 0.345 & $-0.02$ \\
Motion features ($8{\times}32{\times}32$ + diffs) & 0.503 & 0.327 & 0.498 & 0.413 & 0.42 \\
DINOv2 frozen, linear head ($4{\times}224$ px) & 0.523 & 0.345 & 0.531 & 0.460 & 0.64 \\
DINOv2 frozen, boosted head ($4{\times}224$ px) & 0.545 & 0.341 & 0.545 & 0.506 & 0.80~\checkmark \\
3D-conv net, end to end ($8{\times}48{\times}48$) & 0.531 & 0.338 & 0.537 & 0.494 & 0.80~\checkmark \\
\bottomrule
\end{tabular}
\par\smallskip
{\tablefont Majority-class baseline 0.347. \emph{Shuffle}: retrain on scrambled labels; must sit at the baseline. \emph{Tint} / \emph{Azimuth}: held-out scene appearance / camera angle. \emph{Retained}: share of the above-baseline margin surviving the held-out angle; pass (\checkmark) above 0.75; negative means held-out accuracy fell to the baseline.\par}

\end{table*}
\begin{table}[tbp]
\centering
\caption{Refusal rates for the two Cosmos models under the option-position sweep: three phrasings (v1--v3) of the ``cannot be determined'' option, each in last and first list position. Grasp refusal is prompt-robust in both generations; detection refusal collapses for the 8B but barely moves for the Nano.}
\label{tab:sweep-cosmos}
\centering
\small
\setlength{\tabcolsep}{3.5pt}
\begin{tabular}{llcc}
\toprule
Model & Fam. & Last (v1/v2/v3) & First (v1/v2/v3) \\
\midrule
Reason2-8B & A & 100 / 100 / 100 & 100 / 100 / 100 \\
Cosmos3-Nano & A & 100 / 100 / 100 & 97 / 100 / 100 \\
Reason2-8B & B & 94 / 89 / 97 & 3 / 6 / 25 \\
Cosmos3-Nano & B & 100 / 100 / 97 & 92 / 61 / 83 \\
\bottomrule
\end{tabular}
\par\smallskip
{\tablefont Values are refusal rates in \% with the refusal option in last or first list position. Families: A grasp, B detection.\par}

\end{table}
\begin{table*}[tbp]
\centering
\scriptsize
\caption{Regret against the best fixed policy, by prompt version (plain, costs-given, rule-given). The reference asks-all while the model's forced-answer accuracy $a$ is below $p^\star(\Ca)$ and acts-all otherwise; with $a\approx0.32$ on grasp failures the break-even is $\Ca\approx5.3$, so $\Ca{=}6$ is act-optimal. Regret is mean realized cost minus the reference's, $\pm$ a half-width from the 10k-resample bootstrap. Cells marked $\dagger$ are indeterminate: the sign of regret changes within the accuracy estimate's confidence interval. All negative cells are $\dagger$-marked and within their intervals: the reference's accuracy input comes from 25 calibration episodes, so it is optimal in expectation given that estimate, not an oracle, and can be beaten by sampling noise. Overlapping intervals mean condition differences at $n{=}35$ are mostly not separable. Regimes $\Ca\in\{3,10\}$ side by side; the full sweep over $\Ca\in\{1,3,6,10\}$ shows the same pattern.}
\label{tab:regret}
\centering
\begin{tabular}{@{}lllrrr@{}}
\toprule
\multicolumn{6}{c}{$\Ca=3$}\\
\cmidrule(lr){1-6}
Model & Fam. & Cond. & Ask & Cost & Reg. \\
\midrule
Cosmos-Reason2-8B & A & costs & 86\% & 6.71 & $+1.19\,{\scriptscriptstyle\pm1.5}$ \\
Cosmos-Reason2-8B & A & plain & 46\% & 7.23 & $+1.71\,{\scriptscriptstyle\pm1.6}$ \\
Cosmos-Reason2-8B & A & rule & 11\% & 7.63 & $+2.11\,{\scriptscriptstyle\pm1.6}$ \\
Cosmos-Reason2-8B & B & costs & 61\% & 7.56 & $+2.16\,{\scriptscriptstyle\pm1.5}$ \\
Cosmos-Reason2-8B & B & plain & 3\% & 8.31 & $+2.91\,{\scriptscriptstyle\pm1.4}$ \\
Cosmos-Reason2-8B & B & rule & 14\% & 8.64 & $+3.24\,{\scriptscriptstyle\pm1.3}$ \\
Cosmos3-Nano & A & costs & 100\% & 5.43 & $-0.09\,{\scriptscriptstyle\pm1.1}$$^{\dagger}$ \\
Cosmos3-Nano & A & plain & 94\% & 5.83 & $+0.31\,{\scriptscriptstyle\pm1.4}$ \\
Cosmos3-Nano & A & rule & 100\% & 6.00 & $+0.48\,{\scriptscriptstyle\pm1.3}$ \\
Cosmos3-Nano & B & costs & 100\% & 5.67 & $+0.27\,{\scriptscriptstyle\pm1.2}$ \\
Cosmos3-Nano & B & plain & 100\% & 5.67 & $+0.27\,{\scriptscriptstyle\pm1.2}$ \\
Cosmos3-Nano & B & rule & 94\% & 6.61 & $+1.21\,{\scriptscriptstyle\pm1.4}$ \\
InternVL3-8B & A & costs & 86\% & 8.14 & $+2.62\,{\scriptscriptstyle\pm1.5}$ \\
InternVL3-8B & A & rule & 43\% & 8.00 & $+2.48\,{\scriptscriptstyle\pm1.6}$ \\
InternVL3-8B & B & costs & 6\% & 7.00 & $+1.30\,{\scriptscriptstyle\pm1.6}$ \\
InternVL3-8B & B & rule & 3\% & 6.92 & $+1.22\,{\scriptscriptstyle\pm1.6}$ \\
Llama-3.2-11B-V & A & costs & 23\% & 7.40 & $+1.88\,{\scriptscriptstyle\pm1.5}$ \\
Llama-3.2-11B-V & A & plain & 9\% & 8.40 & $+2.88\,{\scriptscriptstyle\pm1.5}$ \\
Llama-3.2-11B-V & A & rule & 3\% & 7.94 & $+2.42\,{\scriptscriptstyle\pm1.5}$ \\
Llama-3.2-11B-V & B & costs & 28\% & 8.50 & $+2.86\,{\scriptscriptstyle\pm1.5}$ \\
Llama-3.2-11B-V & B & plain & 19\% & 7.42 & $+1.78\,{\scriptscriptstyle\pm1.4}$ \\
Llama-3.2-11B-V & B & rule & 8\% & 8.47 & $+2.83\,{\scriptscriptstyle\pm1.3}$ \\
Pixtral-12B & A & costs & 100\% & 9.14 & $+3.56\,{\scriptscriptstyle\pm1.7}$ \\
Pixtral-12B & A & rule & 100\% & 7.14 & $+1.56\,{\scriptscriptstyle\pm1.4}$ \\
Pixtral-12B & B & costs & 100\% & 5.94 & $+0.18\,{\scriptscriptstyle\pm1.2}$ \\
Pixtral-12B & B & rule & 100\% & 7.89 & $+2.13\,{\scriptscriptstyle\pm1.7}$ \\
Qwen2.5-VL-7B & A & costs & 100\% & 6.86 & $+1.34\,{\scriptscriptstyle\pm1.4}$ \\
Qwen2.5-VL-7B & A & plain & 100\% & 5.43 & $-0.09\,{\scriptscriptstyle\pm1.1}$$^{\dagger}$ \\
Qwen2.5-VL-7B & A & rule & 100\% & 6.86 & $+1.34\,{\scriptscriptstyle\pm1.4}$ \\
Qwen2.5-VL-7B & B & costs & 100\% & 6.22 & $+0.46\,{\scriptscriptstyle\pm1.4}$ \\
Qwen2.5-VL-7B & B & plain & 100\% & 5.94 & $+0.18\,{\scriptscriptstyle\pm1.2}$ \\
Qwen2.5-VL-7B & B & rule & 100\% & 6.22 & $+0.46\,{\scriptscriptstyle\pm1.4}$ \\
\bottomrule
\end{tabular}
\hfill
\begin{tabular}{@{}lllrrr@{}}
\toprule
\multicolumn{6}{c}{$\Ca=10$}\\
\cmidrule(lr){1-6}
Model & Fam. & Cond. & Ask & Cost & Reg. \\
\midrule
Cosmos-Reason2-8B & A & costs & 86\% & 13.00 & $+5.20\,{\scriptscriptstyle\pm1.7}$ \\
Cosmos-Reason2-8B & A & plain & 46\% & 10.43 & $+2.63\,{\scriptscriptstyle\pm2.0}$ \\
Cosmos-Reason2-8B & A & rule & 11\% & 8.43 & $+0.63\,{\scriptscriptstyle\pm1.7}$$^{\dagger}$ \\
Cosmos-Reason2-8B & B & costs & 58\% & 11.83 & $+4.83\,{\scriptscriptstyle\pm1.7}$ \\
Cosmos-Reason2-8B & B & plain & 3\% & 8.50 & $+1.50\,{\scriptscriptstyle\pm1.4}$$^{\dagger}$ \\
Cosmos-Reason2-8B & B & rule & 14\% & 9.61 & $+2.61\,{\scriptscriptstyle\pm1.1}$ \\
Cosmos3-Nano & A & costs & 100\% & 13.00 & $+5.20\,{\scriptscriptstyle\pm1.3}$ \\
Cosmos3-Nano & A & plain & 94\% & 12.43 & $+4.63\,{\scriptscriptstyle\pm1.7}$ \\
Cosmos3-Nano & A & rule & 100\% & 13.00 & $+5.20\,{\scriptscriptstyle\pm1.3}$ \\
Cosmos3-Nano & B & costs & 100\% & 12.94 & $+5.94\,{\scriptscriptstyle\pm1.2}$ \\
Cosmos3-Nano & B & plain & 100\% & 12.67 & $+5.67\,{\scriptscriptstyle\pm1.2}$ \\
Cosmos3-Nano & B & rule & 94\% & 13.22 & $+6.22\,{\scriptscriptstyle\pm1.4}$ \\
InternVL3-8B & A & costs & 77\% & 12.71 & $+4.91\,{\scriptscriptstyle\pm1.7}$ \\
InternVL3-8B & A & rule & 43\% & 11.00 & $+3.20\,{\scriptscriptstyle\pm2.0}$ \\
InternVL3-8B & B & costs & 0\% & 7.39 & $-1.61\,{\scriptscriptstyle\pm1.7}$$^{\dagger}$ \\
InternVL3-8B & B & rule & 3\% & 7.11 & $-1.89\,{\scriptscriptstyle\pm1.7}$$^{\dagger}$ \\
Llama-3.2-11B-V & A & costs & 29\% & 9.00 & $+1.20\,{\scriptscriptstyle\pm1.6}$$^{\dagger}$ \\
Llama-3.2-11B-V & A & plain & 9\% & 9.00 & $+1.20\,{\scriptscriptstyle\pm1.6}$$^{\dagger}$ \\
Llama-3.2-11B-V & A & rule & 3\% & 8.14 & $+0.34\,{\scriptscriptstyle\pm1.7}$$^{\dagger}$ \\
Llama-3.2-11B-V & B & costs & 22\% & 10.44 & $+1.84\,{\scriptscriptstyle\pm1.4}$ \\
Llama-3.2-11B-V & B & plain & 19\% & 8.78 & $+0.18\,{\scriptscriptstyle\pm1.4}$$^{\dagger}$ \\
Llama-3.2-11B-V & B & rule & 8\% & 9.06 & $+0.46\,{\scriptscriptstyle\pm1.2}$$^{\dagger}$ \\
Pixtral-12B & A & costs & 100\% & 16.71 & $+8.51\,{\scriptscriptstyle\pm1.7}$ \\
Pixtral-12B & A & rule & 100\% & 14.14 & $+5.94\,{\scriptscriptstyle\pm1.4}$ \\
Pixtral-12B & B & costs & 100\% & 12.94 & $+3.54\,{\scriptscriptstyle\pm1.2}$ \\
Pixtral-12B & B & rule & 100\% & 14.89 & $+5.49\,{\scriptscriptstyle\pm1.7}$ \\
Qwen2.5-VL-7B & A & costs & 100\% & 14.14 & $+6.34\,{\scriptscriptstyle\pm1.6}$ \\
Qwen2.5-VL-7B & A & plain & 100\% & 12.43 & $+4.63\,{\scriptscriptstyle\pm1.1}$ \\
Qwen2.5-VL-7B & A & rule & 100\% & 13.86 & $+6.06\,{\scriptscriptstyle\pm1.4}$ \\
Qwen2.5-VL-7B & B & costs & 100\% & 13.22 & $+3.82\,{\scriptscriptstyle\pm1.4}$ \\
Qwen2.5-VL-7B & B & plain & 100\% & 12.94 & $+3.54\,{\scriptscriptstyle\pm1.2}$ \\
Qwen2.5-VL-7B & B & rule & 100\% & 13.22 & $+3.82\,{\scriptscriptstyle\pm1.4}$ \\
\bottomrule
\end{tabular}
\end{table*}

\end{document}